%% file: romancy_chablat_wenger.tex
\documentclass[a4paper,12pt]{article}
\usepackage[dvips]{graphicx}
\usepackage{theorem}
\input{macro.tex}

\renewcommand{\baselinestretch}{1.24}
\theoremstyle{plain}

\newtheorem{Def}{Definition}

\font\tencmr=cmr10
\font\douzecmr=cmr12

\makeatletter
\def\@normalsize{\@setsize\normalsize{12pt}\xpt\@xpt
\abovedisplayskip 6pt plus 1pt minus 5pt
\belowdisplayskip \abovedisplayskip
\abovedisplayshortskip \z@ plus 3pt
\belowdisplayshortskip 6pt plus 1pt minus 3pt
\let\@listi\@listI}
\def\subsize{\@setsize\subsize{12pt}\xipt\@xipt}
\def\section{\@startsection {section}{1}{\z@}{12pt plus 1pt minus 2pt}
{12pt plus 1pt minus 2pt}{\large\bf}}
\def\subsection{\@startsection {subsection}{2}{\z@}{12pt plus 1pt minus 2pt}
{12pt plus 1pt minus 2pt}{\bf}}
\makeatother
\begin{document}
\date{}
\title {\Large\bf MOVEABILITY AND COLLISION ANALYSIS FOR FULLY-PARALLEL MANIPULATORS}
\author{\begin{tabular}[t]{c}
        Damien Chablat \hfill  Philippe Wenger \\
        {\douzecmr Institut de Recherche en Cybern\'etique de Nantes} \\
        {\douzecmr 1, rue de la No\"e, 44321 Nantes, France} \\
        {\douzecmr Damien.Chablat@lan.ec-nantes.fr} \hfill
        {\douzecmr Philippe.Wenger@lan.ec-nantes.fr} \\
\end{tabular}}
\renewcommand{\baselinestretch}{1.1}
\maketitle
\renewcommand{\baselinestretch}{1.24}
\thispagestyle{empty}
\subsection*{\centering Abstract}
{\em The aim of this paper is to characterize the moveability of fully-parallel manipulators in the presence of obstacles. Fully parallel manipulators are used in applications where accuracy, stiffness or high speeds and accelerations are required \cite{Merlet:97}. However, one of its main drawbacks is a relatively small workspace compared to the one of serial manipulators. This is due mainly to the existence of potential internal collisions, and the existence of singularities. In this paper, the notion of free aspect is defined which permits to exhibit domains of the workspace and the joint space free of singularity and collision. The main application of this study is the moveability analysis in the workspace of the manipulator as well as path-planning, control and design.}
\par
{\bf Key Words:} Parallel Manipulator, Singularity, Working Modes,
Collisions, Aspect, Moveability, Design.
\section{Introduction}
The aim of this paper is to characterize the moveability of
fully-parallel manipulators in the presence of obstacles. Fully
parallel manipulators are used in applications where accuracy,
stiffness or high speeds and accelerations are required
\cite{Merlet:97}. However, one of its main drawbacks is a
relatively small workspace compared to the one of serial
manipulators. This is due mainly to the existence of potential
internal collisions, i.e. collisions between different bodies of
the manipulator. The most current method to avoid such collisions
is to limit the range of the actuated and passive joints by
constant values. This reduces the workspace more than needed. A
better method is to consider virtual limits by software. By
adjusting these limits in function of the configuration of the
manipulator, a significant part of the workspace would be saved.
Another feature which seriously reduces the workspace of
fully-parallel manipulators is the existence of singularities. It
is well known that two Jacobian matrices appear in the kinematic
relations between the joint-rate and the Cartesian-velocity
vectors, which are called the ``inverse kinematics" and the
``direct kinematics" matrices. The study of these matrices allows
to define the parallel and the serial singularities, respectively
\cite{Gosselin:90}. They appear when two solutions of the direct
kinematics (resp. inverse kinematics) meet. A parallel singularity
generally appears inside the workspace and is very difficult to
cross.
\par
The notion of collision-free space is introduced to take into
account the internal/external collisions. However, its projection
onto the workspace is insufficient to conclude as to the
moveability of the manipulator. To solve this problem, we define
the notion of free aspect for general fully parallel manipulators
in the presence of obstacles.
\par
This study is illustrated with a RR-RRR planar parallel
manipulator.
\section{Preliminaries}
\subsection{The fully parallel manipulators}
\begin{Def}
A fully parallel manipulator is a mechanism that includes as many
elementary kinematic chains as the mobile platform does admit
degrees of freedom. Moreover, every elementary kinematic chain
possesses only one actuated joint (prismatic, pivot or kneecap).
Besides, no segment of an elementary kinematic chain can be linked
to more than two bodies \cite{Merlet:97}.
\label{Definition:Fully_Parallel_Manipulator}
\end{Def}
In this study, kinematic chains, or legs \cite{Angeles:97}, are
always independent.
\subsection{Kinematic equations}
The vector of input variables \negr q and the vector of output
variables \negr X for a n-DOF fully parallel manipulator are
related through a system of non linear algebraic equations which
can be written as
\be
   F(\negr X, \negr q)= \negr 0
   \label{equation:system}
\ee
where $\negr 0$ means here the n-dimensional zero vector. \negr q
is the vector of actuated joint variables: $\negr q \in Q$, where
$Q$ is referred to as the joint space of the manipulator. \negr X
is the vector of position and orientation of the moving platform:
$\negr X
\in W$ where $W$ is the workspace of the manipulator. The position
and orientation of all the bodies of the manipulator are fully
defined by (\negr X, \negr q) which will be referred to as {\em
mechanism configuration} in this paper. Differentiating
(\ref{equation:system}) with respect to time leads to the velocity
model
\beqa
   \negr A \negr t + \negr B \negr {\dot q}= \negr 0 \nonumber
\eeqa
The parallel singularities (resp. serial singularities) occur when
the determinant of direct kinematic matrix \negr A (resp.
\negr B) vanishes \cite{Chablat:98}. One can remark that for fully parallel
manipulators, \negr B is always diagonal \cite{Chablat:98}.
\subsection{Working modes}
The {\em working modes} were defined in \cite{Chablat:98} for n-DOF
fully parallel manipulators as follows:
\par
A {\em working mode}, denoted $Mf_i$, is the set of mechanism
configurations for which the sign of $\negr B_{jj}(j = 1~to~n)$
does not change and $\negr B_{jj}$ does not vanish.
\beqa
Mf_i = \left\{ (\negr X, \negr q) \in W \cdot Q \setminus
              \begin{array}{c}
                  sign(\negr B_{jj})=constant~
                  for~(j=1~to~n) \\
                  and~det(\negr B) \neq 0
              \end{array}
       \right\} \nonumber
\eeqa
where $W \cdot Q$ means the cartesian product of $W$ by $Q$.
\par
Therefore, the set of working modes $(Mf=\left\{Mf_i\right\}, i
\in I)$ is obtained while using all permutations of sign of each
term $\negr B_{jj}$.
\subsection{The manipulator and the environment}
\begin{figure}[hbt]
    \begin{center}
    \begin{tabular}{ccc}
       \begin{minipage}[t]{60 mm}
           \centerline{\hbox{\includegraphics[width= 35mm,height= 35mm]{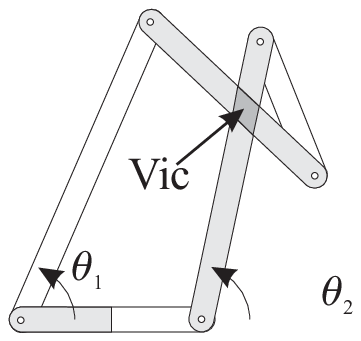}}}
           \caption{Example of internal collision}
           \protect\label{figure:internal_collision}
       \end{minipage} &
       \begin{minipage}[t]{60 mm}
           \centerline{\hbox{\includegraphics[width= 50mm,height= 35mm]{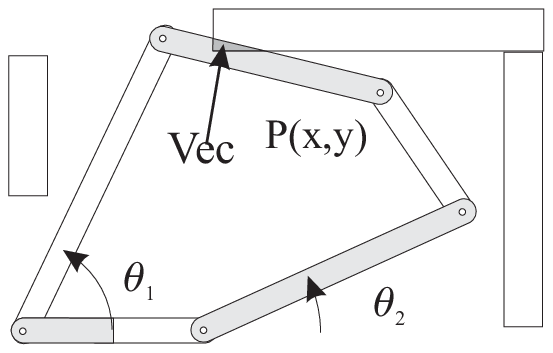}}}
           \caption{Example of external collsion}
           \label{figure:external_collision}
       \end{minipage} &
       \begin{minipage}[t]{30 mm}
           \centerline{\hbox{\includegraphics[width= 30mm,height= 25mm]{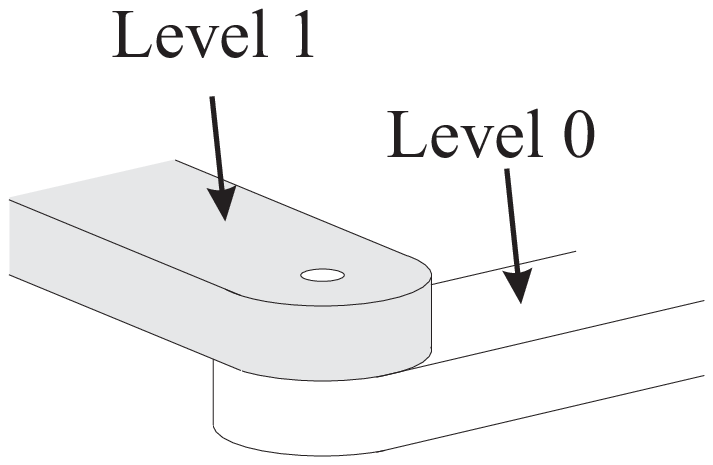}}}
           \caption{Example of revolute}
           \label{figure:pivot_mecanisme}
       \end{minipage}
    \end{tabular}
    \end{center}
\end{figure}

\begin{Def}
Let $\negr {V_M}(\negr  X, \negr q)$, the volume of the fully
parallel manipulator in the mechanism configuration $(\negr X,\negr
q)$.
\end{Def}
\beqa
\negr {V_M}(\negr X, \negr q) = \negr b \cup
                               (\cup_{k=0,n\times m - 1} \negr c_{k}(\negr X, \negr q)) \cup
                               \negr {pl}(\negr X) \nonumber
\eeqa
where
\begin{itemize}
\item $\negr b$ is the volume of the fixed base of the fully parallel manipulator;
\item $\negr {pl}(\negr X)$ is the volume of the mobile platform of the fully parallel
manipulator in the platform configuration (\negr X);
\item  $\negr c_{k}(\negr X,\negr q)$ with $k= m \times i + j$ is the volume of the link $j$
of the leg $i$ where $n$ is the number of legs and $m$ the number
of links of the leg $i$ of the fully parallel manipulator.
\end{itemize}
\begin{Def}
Let $\negr {V_{ic}}(\negr X, \negr q)$, {\em the volume of the
internal collisions}, i.e. the set of all the volumes in collision
between the links of the manipulator in the mechanism configuration
($\negr X, \negr q)$ (Figure~\ref{figure:internal_collision}):
\beqa
\negr {V_{ic}}(\negr X, \negr q) =
  \begin{array}{l}
    (\cup_{i= 0, n \times m - 1} (\negr c_{i} \cap \negr b)) ~ \cup ~
    (\cup_{i= 0, n \times m - 1} (\negr c_{i} \cap \negr {pl}(\negr X))) ~ \cup ~\\
    (\cup_{i= 0, n \times m - 1} (\cup_{j = i + 1, n \times m - 1} (\negr c_{i} \cap \negr c_{j})))
  \end{array} \nonumber
\eeqa
\end{Def}
\begin{Def}
Let $\negr{V_{ec}}$, {\em the volume of the external collisions},
i.e. the set of all the volumes in collision between the mechanism
and the obstacles (Figure~\ref{figure:external_collision}):
\end{Def}
\beqa
   \negr{V_{ec}} = \negr {V_M} \cap (\cup_{s=1, No} \negr {Obst}_s)  \nonumber
\eeqa
where $No$ is the number of obstacles.
\section{The free aspects of fully parallel manipulators}
\subsection{Collision-free space, workspace and joint space}
\begin{Def}
Let $S_{CF}$, the {\em collision-free space} of the fully parallel
manipulator:
\end{Def}
\beqa
S_{CF}=
    \left\{(\negr X, \negr q) \in W \cdot Q \setminus
                \negr{V_{ec}}(\negr X, \negr q) =\emptyset ~and~
                \negr{V_{ic}}(\negr X, \negr q) = \emptyset
    \right\} \nonumber
\eeqa
A free mechanism configuration $(\negr X, \negr q)$ is a mechanism
configuration for which the manipulator is free of internal and
external collisions.
\begin{Def}
The projection ${\bf \Pi}_W$ of the collision-free space $S_{CF}$
onto the workspace yields the {\em collision-free workspace}
$W_{F}$:
\end{Def}
\beqa
W_{F} = {\bf \Pi}_W(S_{CF}) \nonumber
\eeqa
\begin{Def}
The projection ${\bf \Pi}_Q$ of the collision-free space $S_{CF}$
onto the joint space yields the {\em collision-free joint space}
$Q_{F}$:
\end{Def}
\beqa
Q_{F} = {\bf \Pi}_Q(S_{CF}) \nonumber
\eeqa
\subsection{The free aspect}
The notion of aspect was introduced by \cite{Borrel:86} to cope
with the existence of multiple inverse kinematic solutions in
serial manipulators. Recently, the notion of aspect was defined for
fully parallel manipulators with only one inverse kinematic
solution \cite{Wenger:97} and for fully parallel manipulators with
several inverse and direct kinematic solutions \cite{Chablat:98}.
However, no collision was considered in these last definitions.
\par
In this section, the  notion of free aspect is defined formally for
fully parallel manipulators in the presence of collisions.
\begin{Def}
\label{definition:Aspect}
The free aspects $A_{Fij}$ are defined as the maximal sets in $W
\cdot Q$ so that
\end{Def}
\begin{itemize}
\item $A_{Fij} \subset W \cdot Q$;
\item $A_{Fij}$ is connected.
\item $A_{Fij} = \left\{
                   (\negr X, \negr q) \in Mf_i \setminus
                     det(\negr A) \neq 0 ~and~
                     \negr{V_{ec}}(\negr X, \negr q) = \emptyset ~and~
                     \negr{V_{ic}}(\negr X, \negr q) = \emptyset
                  \right\}$
\end{itemize}
In other words, the free aspects $A_{Fij}$ are the maximal
singularity-free domains of $W \cdot Q$ without neither internal
nor external collisions.
\begin{Def}
The projection ${\bf \Pi}_W$ of the free aspects onto the workspace
yields the free W-aspects $W\!A_{Fij}$:
\end{Def}
\begin{itemize}
\item $W\!A_{Fij} = {\bf \Pi}_W(A_{Fi})$;
\item $W\!A_{Fij}$ is connected.
\end{itemize}
The free W-aspects are the maximal singularity-free domains in the
workspace without neither internal nor external collisions.
\begin{Def}
The projection ${\bf \Pi}_Q$ of the free aspects in the joint space
yields the free Q-aspects $QA_{Fij}$:
\end{Def}
\begin{itemize}
\item $QA_{Fij} = {\bf \Pi}_Q(A_{Fi})$;
\item $QA_{Fij}$ is connected.
\end{itemize}
The free Q-aspects are the maximal singularity-free domains in the
joint space without neither internal nor external collisions.
\section{Applicative example}
\begin{figure}[hbt]
    \begin{center}
      \includegraphics[width= 78mm,height= 40mm]{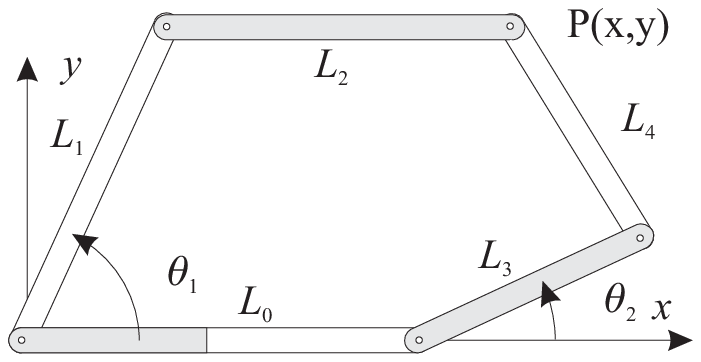}
      \caption{Planar fully parallel manipulator}
      \label{figure:parallel_manipulator}
    \end{center}
\end{figure}

In this section, a planar manipulator is used as illustrative
example. This is a five-bar, revolute ($R$)-closed-loop linkage, as
displayed in figure~\ref{figure:parallel_manipulator}. The actuated
joint variables are $\theta_1$ and $\theta_2$, while the output
variables are the ($x$, $y$) coordinates of the revolute center
$P$. The passive and the active joints will be assumed unlimited in
this study. Lengths $L_0$, $L_1$, $L_2$, $L_3$, and $L_4$ define
completely the geometry of this manipulator. We assume here the
dimensions $L_0=8$, $L_1=7$, $L_2=7$, $L_3=5$ and $L_4=5$, in
certain units of length that we need not specify.
\par
In this example, the environment is assumed free of obstacles so
that only internal collisions are taken into account. The quadtree
model is used to represent the moveability regions in the workspace
and the joint space. They are calculated using discretization and
enrichment techniques \cite{Chablat:96}.
\subsection{Collision Free Workspace and Joint Space}
\begin{figure}[hbt]
    \begin{center}
    \begin{tabular}{cccc}
       \begin{minipage}[t]{50 mm}
           \centerline{\hbox{\includegraphics[width= 35mm,height= 35mm]{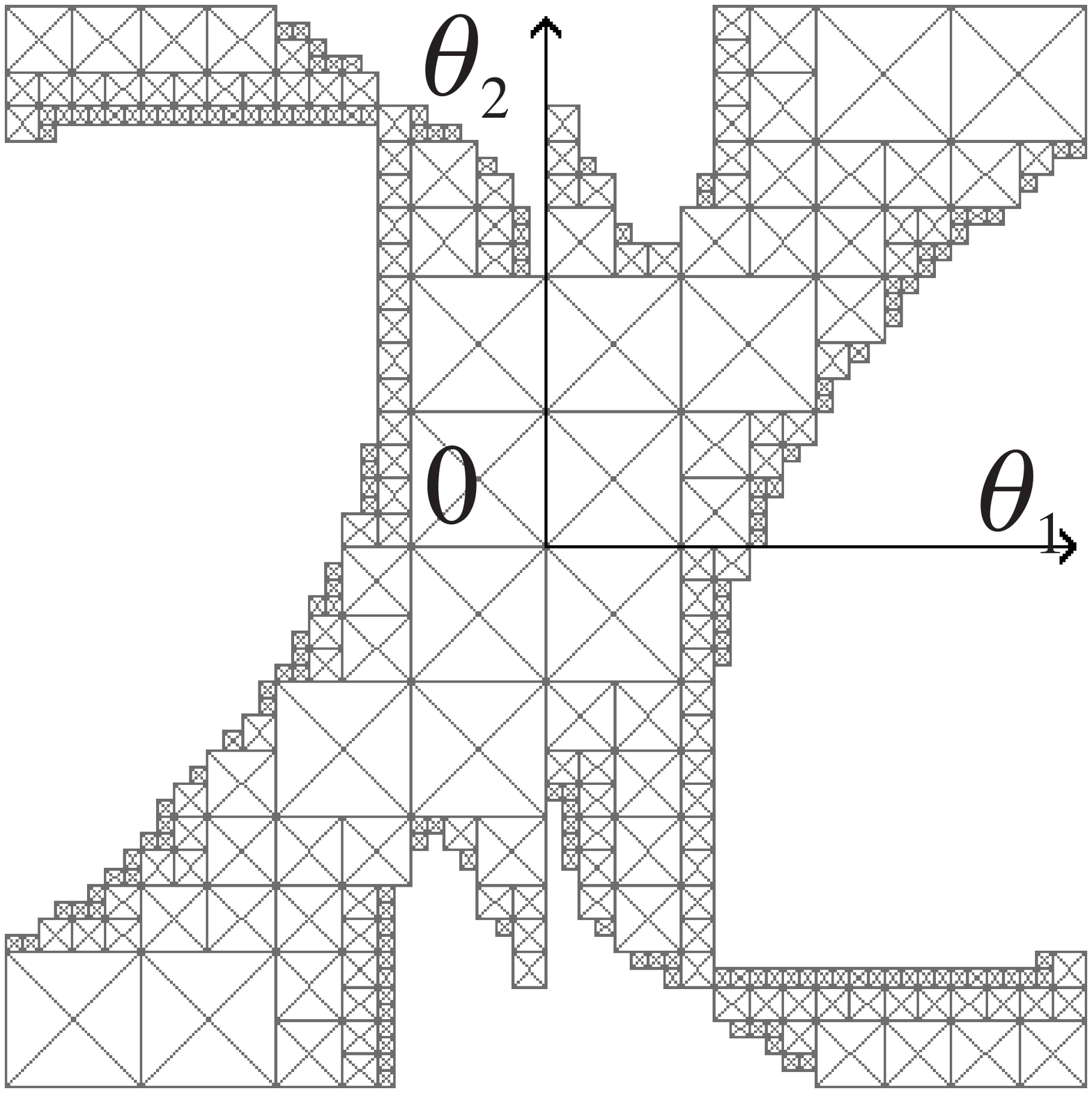}}}
           \caption{Collision free joint space}
           \label{figure:free_joint_space}
       \end{minipage} &
       \begin{minipage}[t]{50 mm}
           \centerline{\hbox{\includegraphics[width= 35mm,height= 35mm]{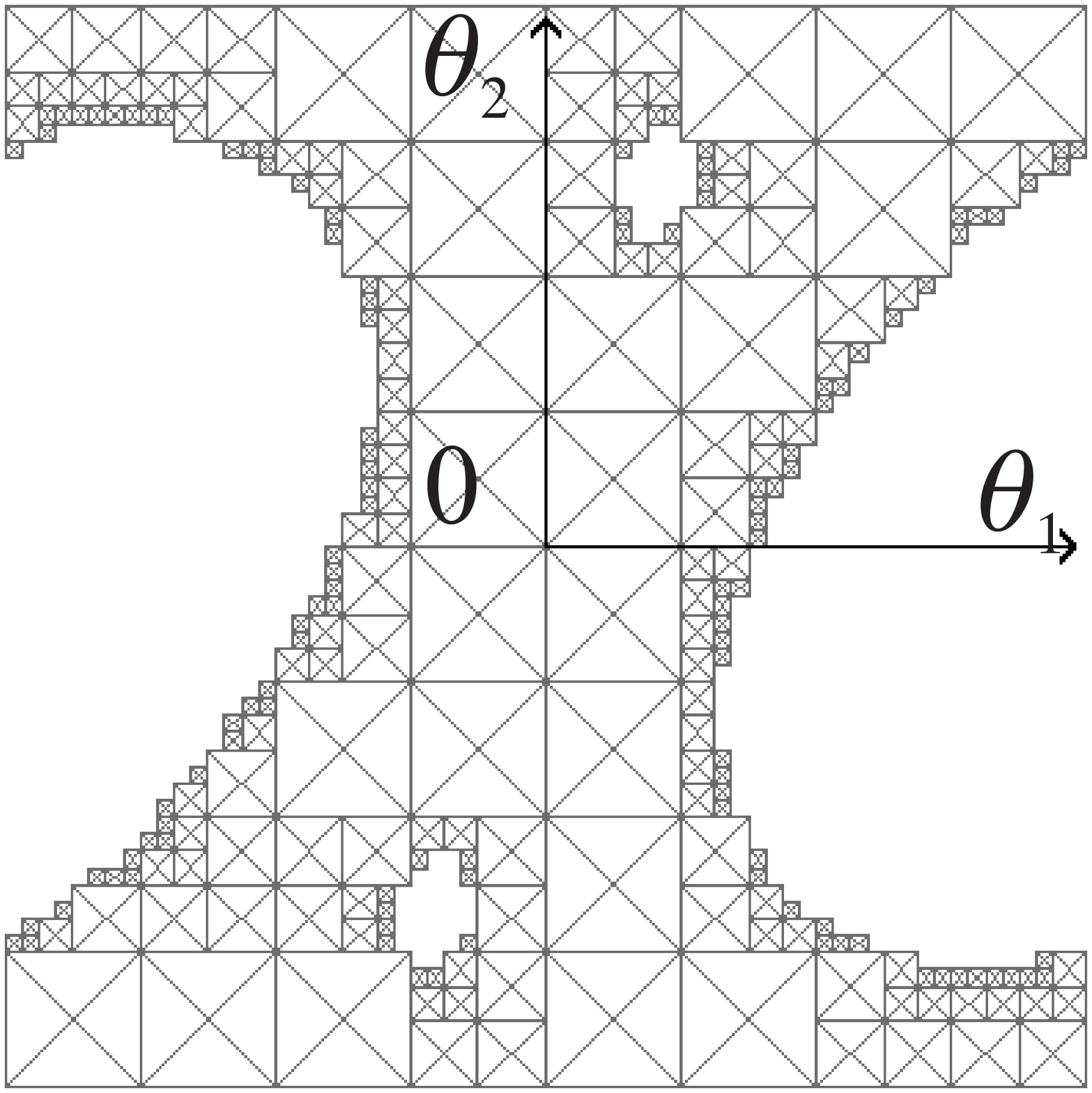}}}
           \caption{Joint space}
           \label{figure:joint_space}
       \end{minipage} &
       \begin{minipage}[t]{50 mm}
           \centerline{\hbox{\includegraphics[width= 35mm,height= 35mm]{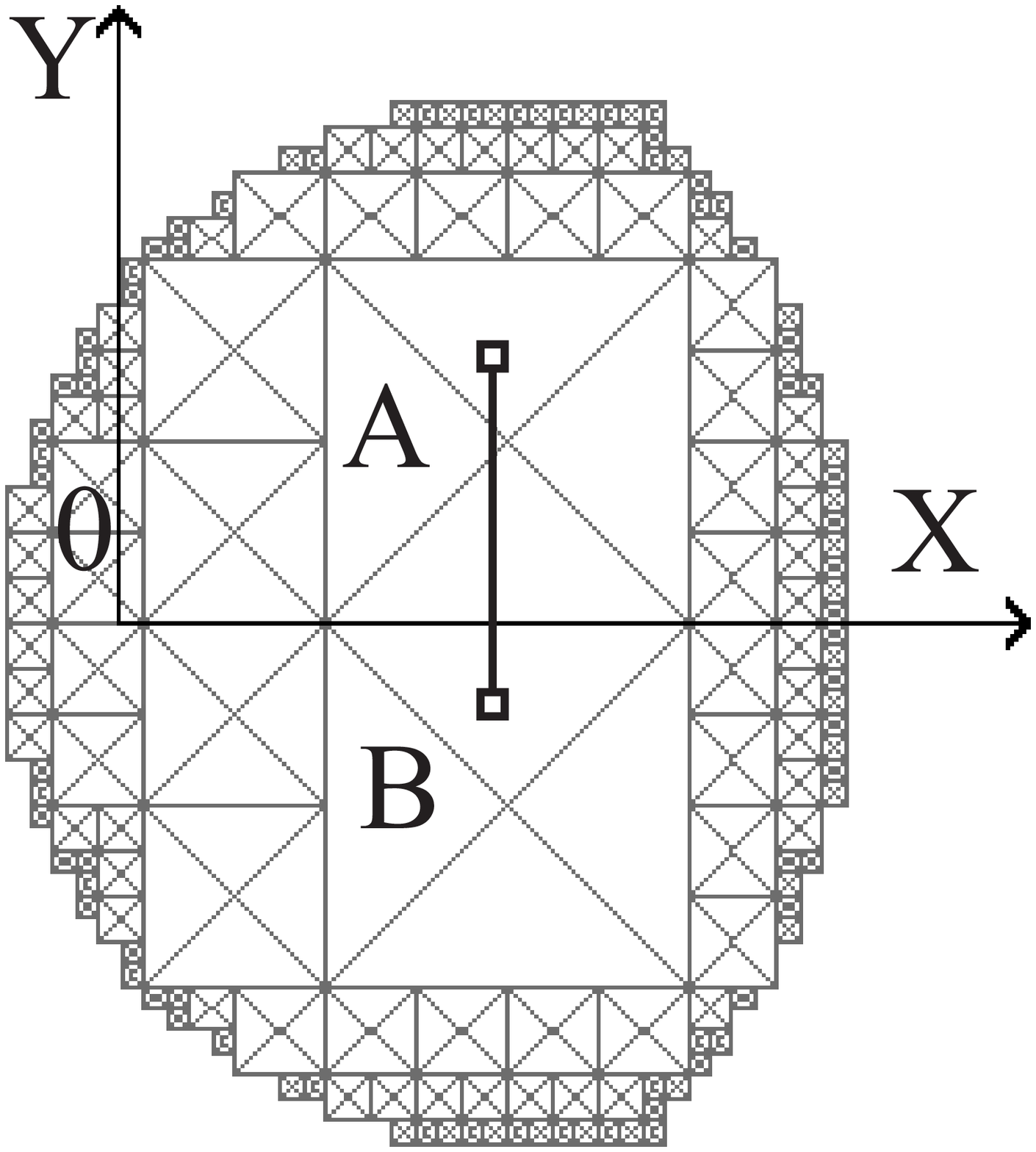}}}
           \caption{Workspace and collision-free workspace}
           \label{figure:workspace}
       \end{minipage}
    \end{tabular}
    \end{center}
\end{figure}

In this example, the collision free joint space (Figure~\ref{figure:free_joint_space}) is smaller than the joint space (Figure~\ref{figure:joint_space}) but the collision-free workspace is similar to the workspace (Figure~\ref{figure:workspace}).
\par
However, in this workspace, any trajectory is not possible. For
example, the trajectory (AB) is infeasible because $P$ goes through
the fixed base $(L_0)$ yielding a collision.
\par
Therefore, the free workspace is insufficient to conclude as to the
moveability of the manipulator. We need to compute the free
aspects.
\subsection{Free aspect}
\begin{figure}[!hbt]
    \begin{center}
        \includegraphics[width= 160mm,height= 30mm]{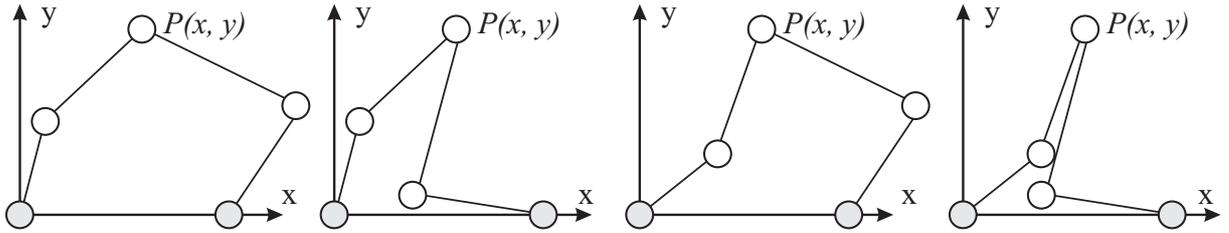}
        \caption{The four working modes of the RR-RRR manipulator}
        \label{figure:working_mode}
    \end{center}
\end{figure}

In this section, the free W-aspects and free Q-aspects are
displayed in the case where $det(A) > 0$ for the 4 existing working
modes (Figure \ref{figure:working_mode}). The other aspects
($det(A) < 0$) are located symmetrically with regard to the $X$
axis.
\par
\begin{figure}[hbt]
    \begin{center}
    \begin{tabular}{cccc}
       \begin{minipage}[t]{35 mm}
           \centerline{\hbox{\includegraphics[width= 35mm,height=35mm]{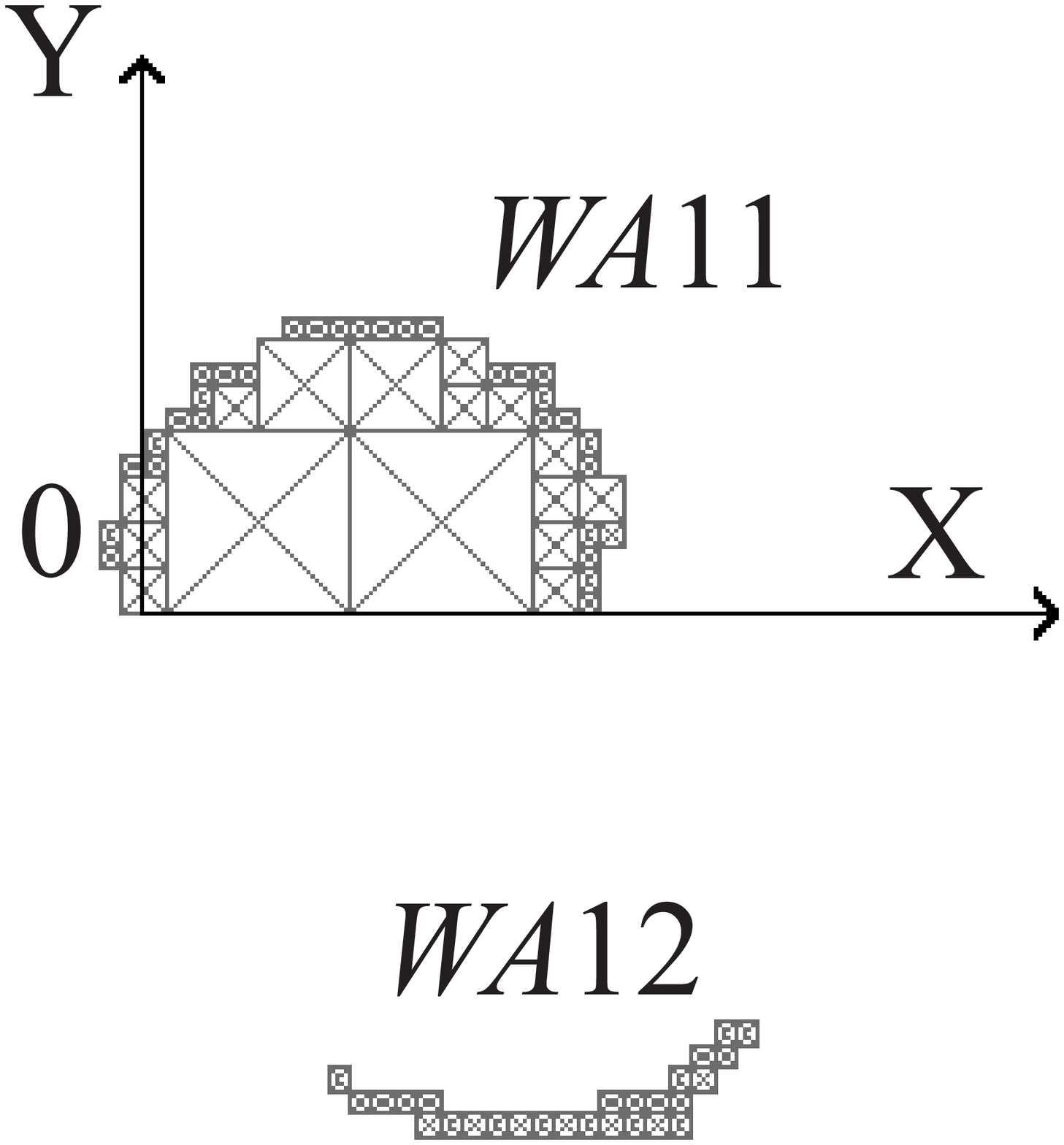}}}
           \caption{Free W-aspects $W\!A_{11}$ and $WA_{12}$}
           \label{figure:aspect_1_2}
       \end{minipage} &
       \begin{minipage}[t]{35 mm}
           \centerline{\hbox{\includegraphics[width= 35mm,height=35mm]{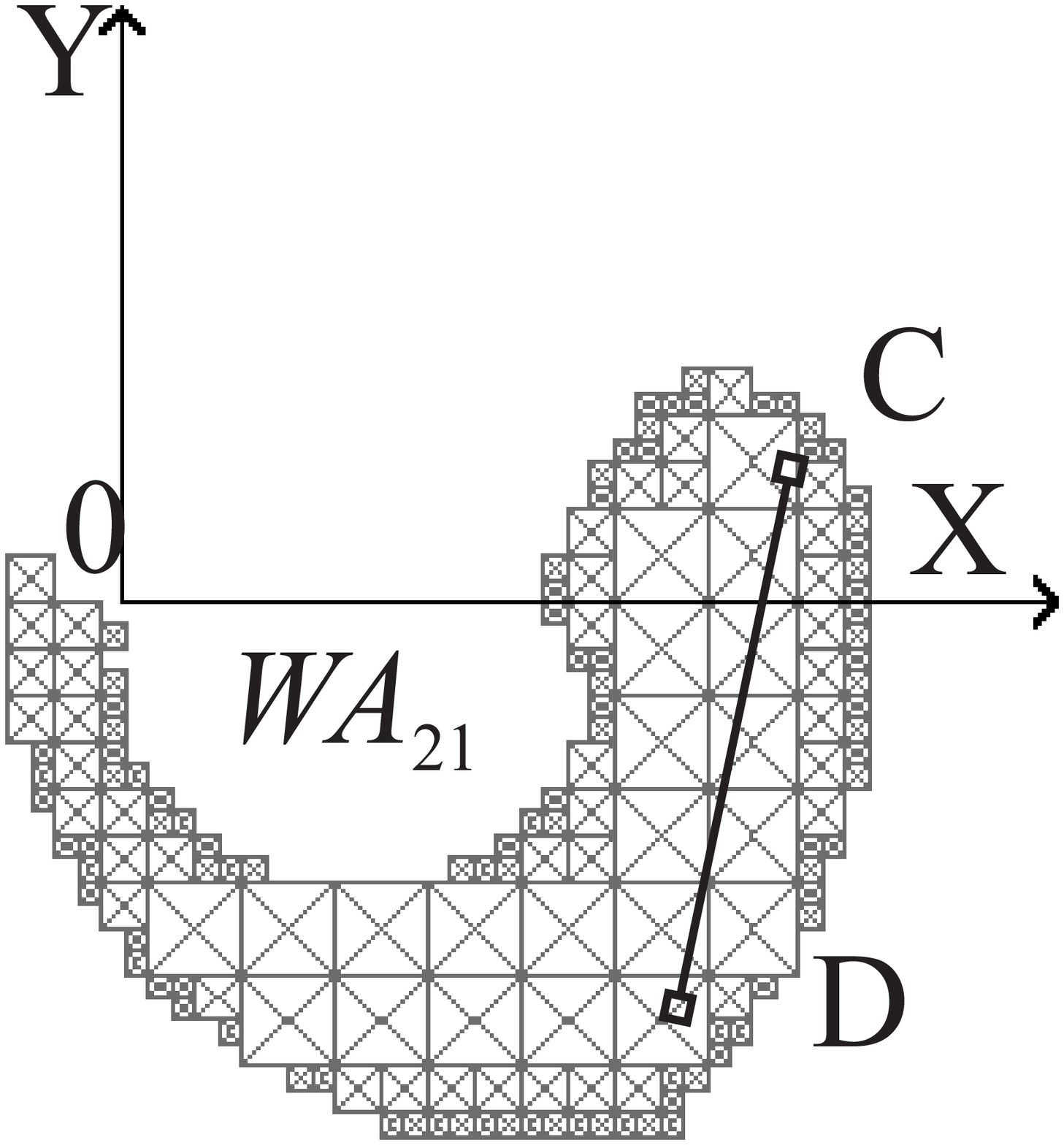}}}
           \caption{Free W-aspect $W\!A_{21}$}
           \label{figure:aspect_3}
       \end{minipage} &
       \begin{minipage}[t]{35 mm}
           \centerline{\hbox{\includegraphics[width= 35mm,height=35mm]{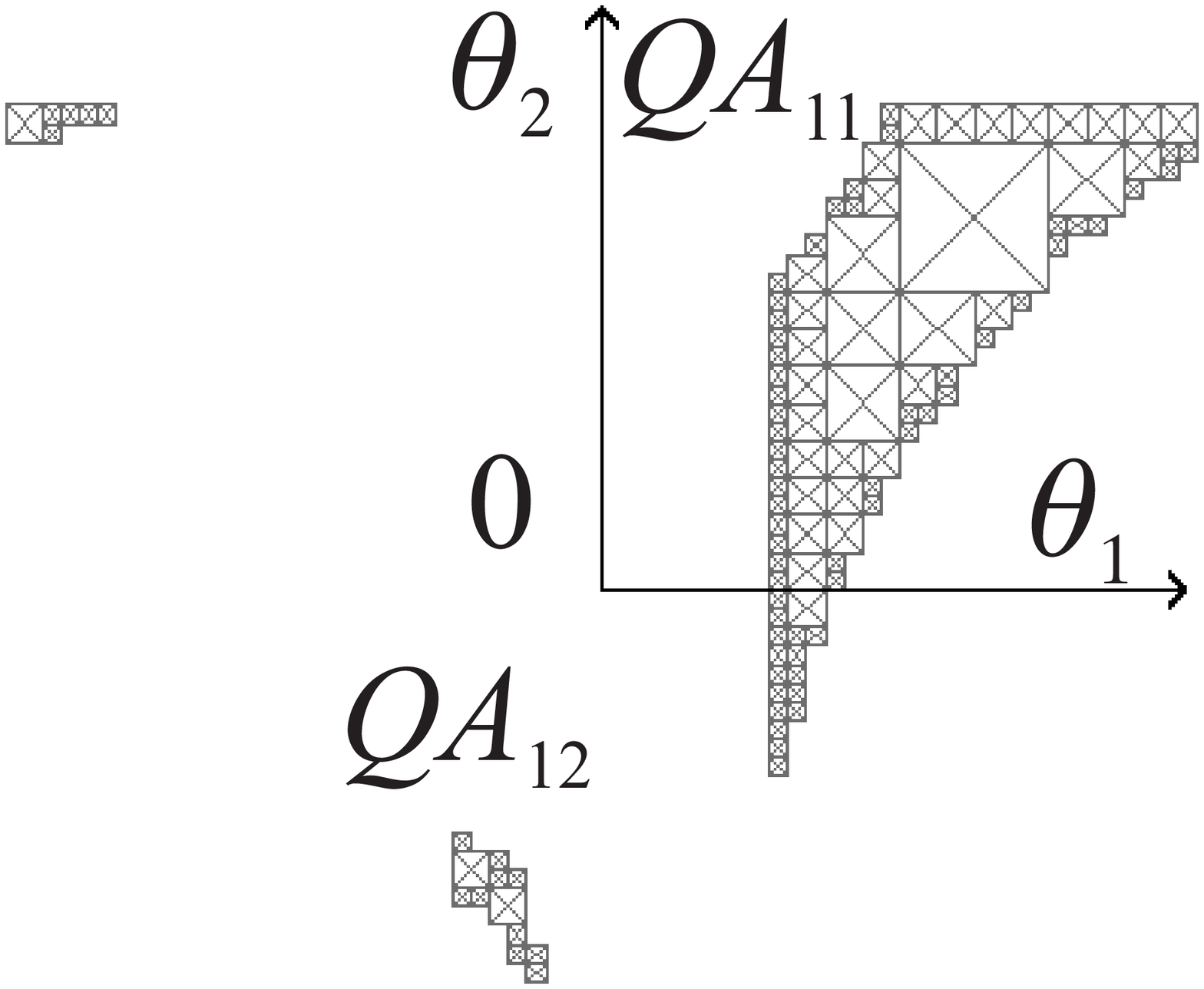}}}
           \caption{Free Q-aspects $QA_{11}$ and $QA_{12}$}
       \end{minipage} &
       \begin{minipage}[t]{35 mm}
           \centerline{\hbox{\includegraphics[width= 35mm,height=35mm]{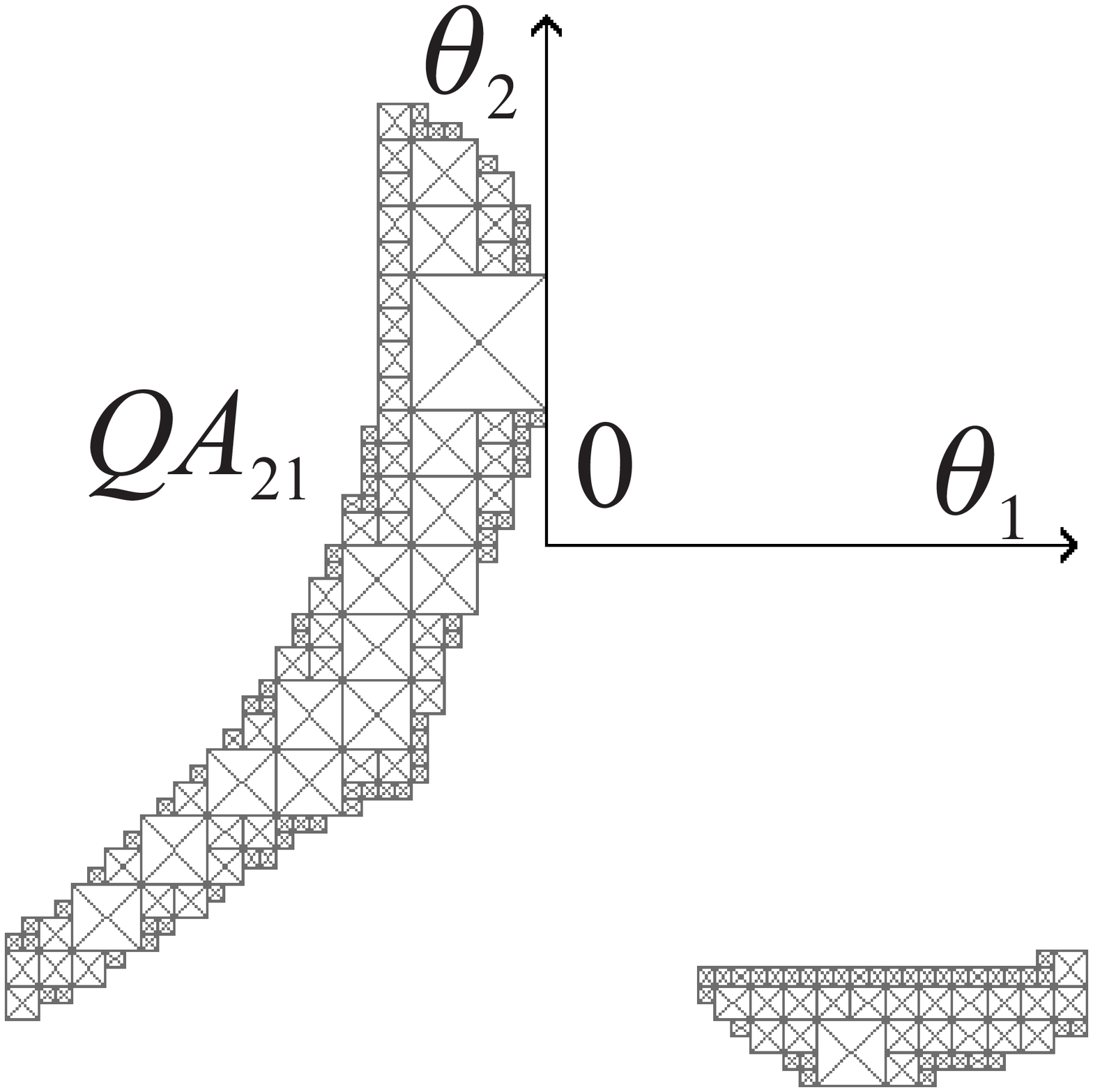}}}
           \caption{Free Q-aspect $QA_{21}$}
       \end{minipage} \\
       \begin{minipage}[t]{160 mm}
           {\hbox{\vsize=0mm}}
       \end{minipage} \\
       \begin{minipage}[t]{35 mm}
           \centerline{\hbox{\includegraphics[width= 35mm,height=35mm]{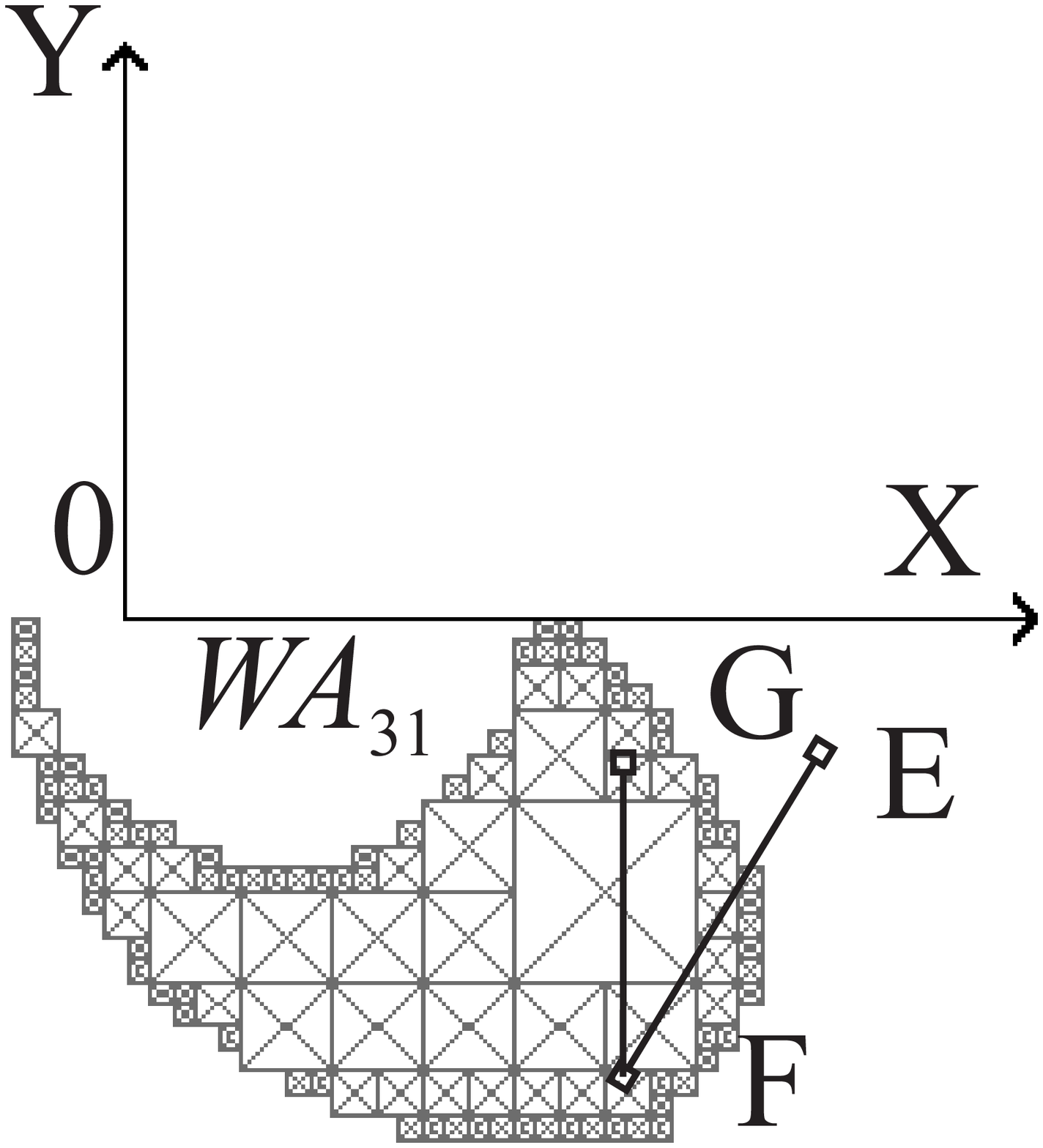}}}
           \caption{Free W-aspect $W\!A_{31}$}
           \label{figure:aspect_4}
       \end{minipage} &
       \begin{minipage}[t]{35 mm}
           \centerline{\hbox{\includegraphics[width= 35mm,height=35mm]{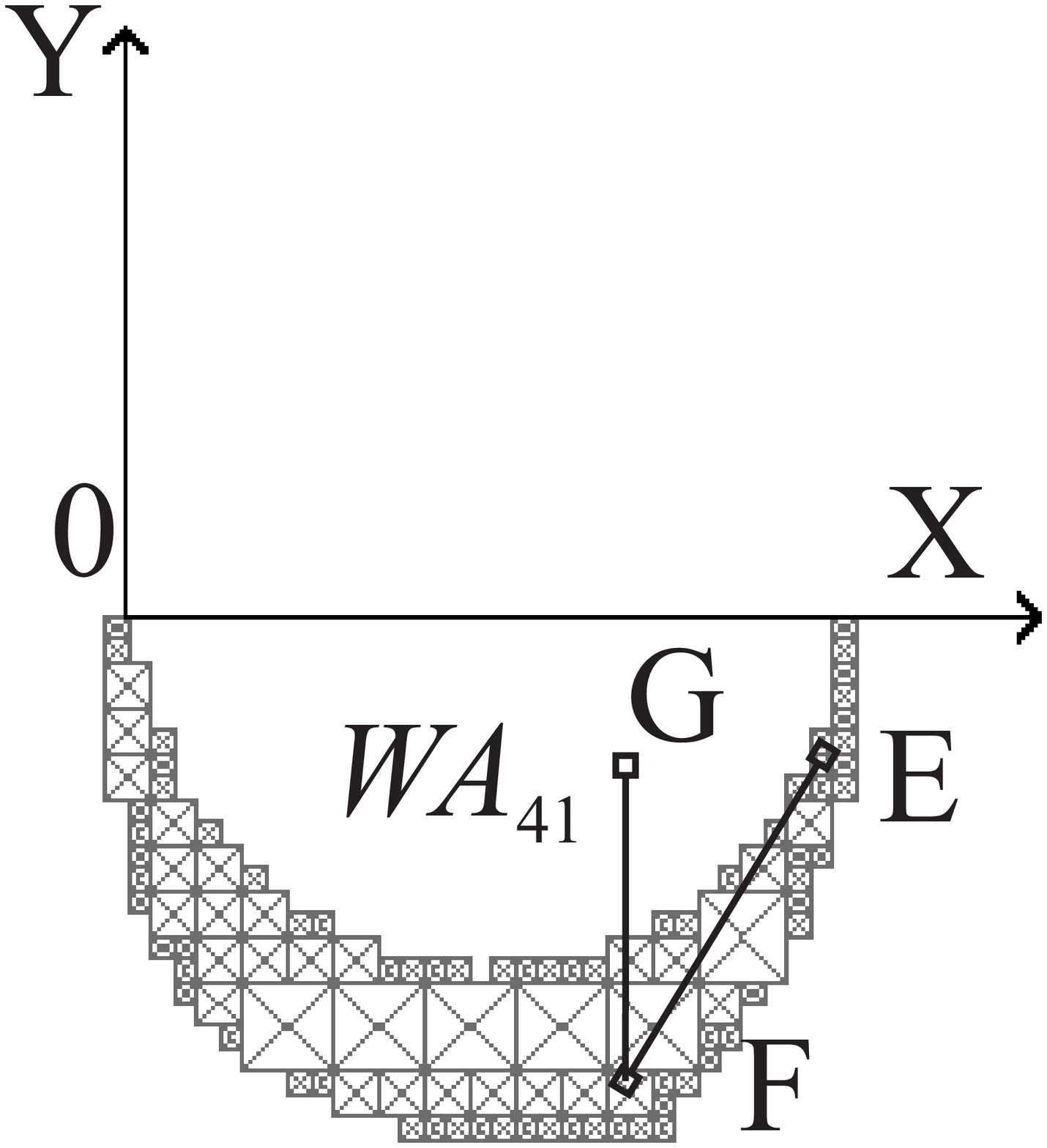}}}
           \caption{Free W-aspect $W\!A_{41}$}
           \label{figure:aspect_5}
       \end{minipage} &
       \begin{minipage}[t]{35 mm}
           \centerline{\hbox{\includegraphics[width= 35mm,height=35mm]{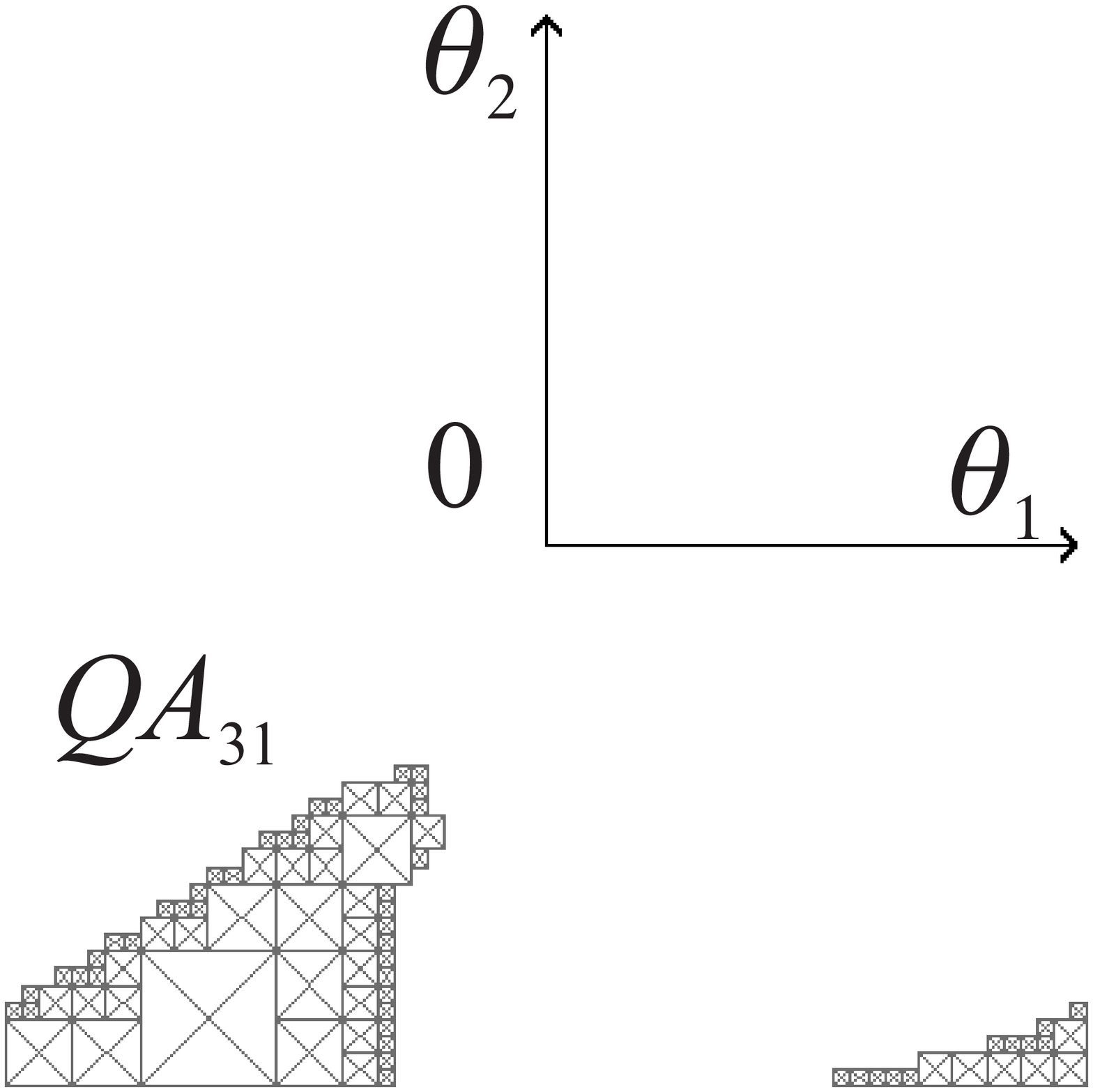}}}
           \caption{Free Q-aspect $Q\!A_{31}$}
       \end{minipage} &
       \begin{minipage}[t]{35 mm}
           \centerline{\hbox{\includegraphics[width= 35mm,height=35mm]{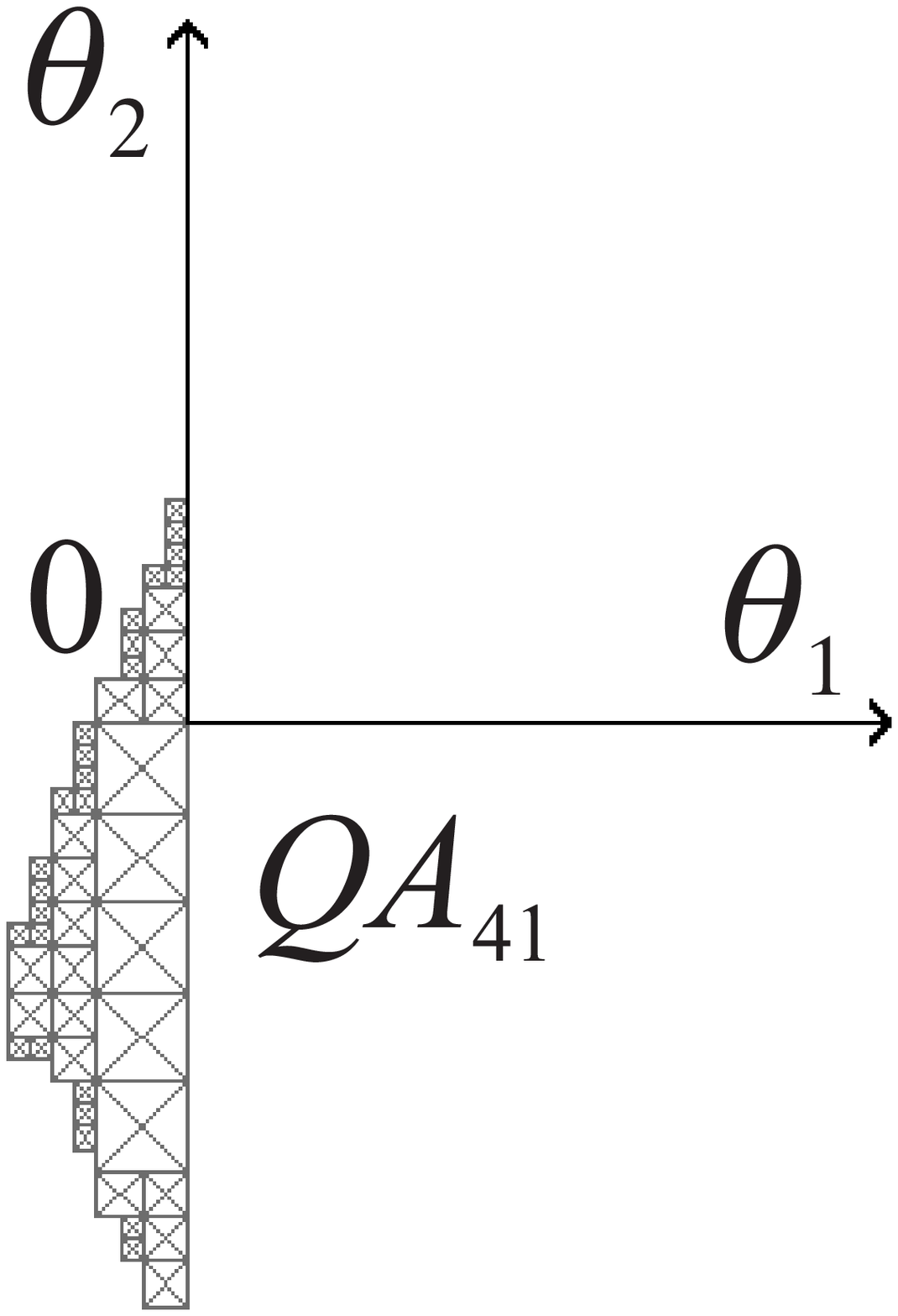}}}
           \caption{Free Q-aspect $Q\!A_{41}$}
       \end{minipage}
    \end{tabular}
    \end{center}
\end{figure}

We can remark that the number of aspects varies according to the
working mode (Figure \ref{figure:aspect_1_2}). The boundary of the
aspects are defined by the parallel and serial singularities as
well as the collisions.
\par
In a free W-aspect, the manipulator can achieve any continuous
trajectory since the free W-aspects are free of collision and
singularity by definition. For instance, the trajectory (CD) is
feasible (Figure \ref{figure:aspect_3}). On the other hand, if a
trajectory is given which does not lie in a same aspect like the
trajectory (EFG) (Figures \ref{figure:aspect_4} and
\ref{figure:aspect_5}), a change of working mode is required. The
change of working mode makes the output link leave the trajectory.
Thus, the continuous trajectory (EFG) is infeasible.
\section{Conclusion}
In this paper, new notions were defined to take into account the
singularities and the presence of internal/external collisions in
the moveability analysis of fully parallel manipulators. To take
into account collisions, the collision free space, the free
workspace and the free joint space were defined. Then, by using the
notion of working modes, a general definition of free aspects was
introduced. The free aspects are defined as the maximal sets of $W
\cdot Q$ which are free of singularities and collisions. Upon
projecting these free aspects onto the workspace (resp. the joint
space), we get the W- aspects (resp. the Q-aspects). The W-aspects
were shown to be the regions of the workspace were any continuous
trajectory is feasible. These sets are of high interest for the
trajectory planning, the control and the design of fully-parallel
manipulators. Further research work is conducted by the authors on
the design of fully-parallel manipulators to minimize the effects
of internal collisions.
\renewcommand{\baselinestretch}{0.6}
\setlength{\bibindent}{0.0cm}
{
\bibliographystyle{unsrt}
\tencmr
}
\end{document}

%% file: macro.tex
\newcommand{\be}{\begin{equation}}
\newcommand{\ee}{\end{equation}}
\newcommand{\beq}{\begin{equation}}
\newcommand{\eeq}{\end{equation}}
\newcommand{\bed}{\begin{displaymath}}
\newcommand{\eed}{\end{displaymath}}
\newcommand{\beqa}{\begin{eqnarray}}
\newcommand{\eeqa}{\end{eqnarray}}
\newcommand{\beqann}{\begin{eqnarray*}}
\newcommand{\eeqann}{\end{eqnarray*}}
\newcommand{\bseq}{\begin{subequation}}
\newcommand{\eseq}{\end{subequation}}

\newcommand{\ba}{\begin{array}}
\newcommand{\ea}{\end{array}}

\newcommand{\negr}[1]{{\bf {#1}}}